\PassOptionsToPackage{table}{xcolor}

\documentclass{article}
\usepackage[utf8]{inputenc}
\usepackage{microtype}
\usepackage{graphicx}
\usepackage{booktabs} 
\usepackage{comment}
\usepackage{arydshln}
\usepackage{makecell}
\usepackage{hyperref}
\usepackage{mathtools}
\usepackage{color}
\usepackage{listings}
\usepackage{xcolor}

\usepackage{graphicx}
\usepackage{CJKutf8}
\usepackage{url}
\usepackage{multicol,multirow}
\usepackage{makecell}
\usepackage{algorithmic}
\usepackage{algorithm}
\usepackage{booktabs}
\usepackage{textcomp}

\usepackage{bm}
\usepackage{parskip}
\usepackage[T1]{fontenc}    
\usepackage{booktabs}       
\usepackage{amsmath, amsthm}
\usepackage{amssymb}
\usepackage{amsfonts}
\usepackage{times}
\usepackage{latexsym}
\usepackage{amsmath, amsthm}
\usepackage{amssymb}
\usepackage{amsfonts}

\usepackage{tabularx}

\usepackage{tabularx}       
\usepackage{verbatim}       
\usepackage{fancyvrb}       
\usepackage{listings}       
\usepackage{enumitem}       
\usepackage{amsmath}        
\usepackage{geometry}       

\usepackage{booktabs}       
\usepackage{float}          
\usepackage{caption}        

\usepackage[most]{tcolorbox}

\usepackage{longtable}
\usepackage{listings}
\usepackage{booktabs, listings, enumitem}
\usepackage{listings}
\usepackage{graphicx}
\usepackage{adjustbox}

\definecolor{promptblue}{RGB}{200, 230, 255}
\definecolor{exemplargreen}{RGB}{220, 255, 220}
\definecolor{protocolyellow}{RGB}{255, 255, 200}

\setlist{nolistsep}
\renewcommand{\arraystretch}{0.8}

\usepackage{booktabs,  tcolorbox, listings, etoolbox, tabularx}
\tcbuselibrary{listings, skins, breakable}

\usepackage{fontawesome5}

\definecolor{taskblue}{RGB}{0,99,177}
\definecolor{refgreen}{RGB}{0,150,85}
\definecolor{subviolet}{RGB}{131,76,190}
\definecolor{templateblue}{RGB}{1, 128, 134}
\definecolor{refgreenDark}{RGB}{0, 90, 45}
\definecolor{analysisblue}{HTML}{1E90FF} 
\definecolor{algoyellow}{HTML}{A0522D}   

\lstdefinestyle{codestyle}{
  language=Python,
  basicstyle=\footnotesize\ttfamily,
  frame=single,
  numbers=left,
  numberstyle=\tiny,
  xleftmargin=1.5em,
  framexleftmargin=1.5em,
  keywordstyle=\color{taskblue},
  commentstyle=\itshape\color{gray},
  stringstyle=\color{orange},
  showstringspaces=false,
  breaklines=true,
  tabsize=2
}
\definecolor{kwblue}{HTML}{005CFF}       
\definecolor{strred}{HTML}{B80034}    
\definecolor{codebg}{RGB}{245,248,250}
\definecolor{argsc}{RGB}{0,128,128}
\definecolor{codegreen}{HTML}{189399}

\lstdefinestyle{py}{
  language        = Python,
  basicstyle      = \tiny\ttfamily,
  numbers         = left,
  numberstyle     = \tiny\color{gray},
  stepnumber      = 1,
  keywordstyle    = \color{kwblue}\bfseries,
  commentstyle    = \color{gray},
  stringstyle     = \color{strred},
  breaklines      = true,
  showstringspaces= false,
  tabsize         = 4,
  backgroundcolor = \color{templateblue!5},
  frame           = single,
  xleftmargin     = 3em,             
  framexleftmargin= 2.5em,  
  rulecolor       = \color{codebg},
  framesep        = 6pt,        
  literate        = {Args}{{\textcolor{argsc}{Args}}}4
}
\usepackage{longtable,booktabs} 
\colorlet{subvioletstrong}{subviolet!80!black} 
\usepackage{listings}
\lstdefinestyle{afteroptimcode}{
  language=Python, numbers=left, frame=none,
  numberstyle   = \tiny,        
  numbersep     = 3pt,          
  xleftmargin   = 2pt,          
  framexleftmargin = 0pt,       
  basicstyle=\ttfamily\footnotesize, keywordstyle=\color{subviolet},
  breaklines=true, columns=fullflexible
}

\usepackage{listings}
\lstdefinestyle{beforeoptimcode}{
  language=Python, numbers=left, frame=none,
  numberstyle   = \tiny,        
  numbersep     = 3pt,          
  xleftmargin   = 2pt,          
  framexleftmargin = 0pt,       
  basicstyle=\ttfamily\footnotesize, keywordstyle=\color{subviolet},
  breaklines=true, columns=fullflexible
}

\usepackage[normalem]{ulem} 
\usepackage{multirow}
\usepackage{nicematrix} 
\usepackage{pgfplots}   
\usepackage{xcolor}     
\usepackage{array}      
\usepackage{arydshln}
\definecolor{code-highlight-blue}{HTML}{2696f0}
\definecolor{code-highlight-green}{HTML}{7eb547}
\definecolor{code-highlight-yellow}{HTML}{fdcc3b}
\definecolor{code-highlight-purple}{HTML}{ab4abb}
\definecolor{code-highlight-red}{HTML}{f3473a}
\definecolor{code-highlight-orange}{HTML}{fe970c}

\DeclareCaptionLabelFormat{figsub}{Figure~\thefigure(\alph{subfigure})}

\usepackage{pgfplots}
\usepackage{subcaption}

\newcommand{\bluecode}[1]{\textcolor{code-highlight-blue}{\bfseries\footnotesize\ttfamily #1}}
\newcommand{\greencode}[1]{\textcolor{code-highlight-green}{\bfseries\footnotesize\ttfamily #1}}
\newcommand{\yellowcode}[1]{\textcolor{code-highlight-yellow}{\bfseries\footnotesize\ttfamily #1}}
\newcommand{\purplecode}[1]{\textcolor{code-highlight-purple}{\bfseries\footnotesize\ttfamily #1}}
\newcommand{\redcode}[1]{\textcolor{code-highlight-red}{\bfseries\footnotesize\ttfamily #1}}

\usepackage{marvosym}

\newcommand{\emailmark}{%
    \textsuperscript{\large\Letter}%
}

\newcommand{\emailtext}[1]{%
    \begingroup
    \renewcommand{\thefootnote}{\large\Letter}%
    \footnotetext{#1}%
    \endgroup
}

\newcommand{\Romnum}[1]{\MakeUppercase{\romannumeral #1}}
\usepackage{wrapfig}

\title{CUDA-L1: Improving CUDA Optimization via  Contrastive  \\ Reinforcement Learning\\}

\author{Xiaoya Li, Albert Wang, Guoyin Wang, Jiwei Li and Chris Shum}

\date{\textbf{\large Ornith Team}\\\vspace{0.1cm}\includegraphics[scale=0.05]{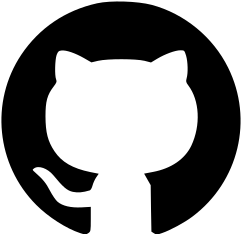} 
\href{https://github.com/ornith-ai/CUDA-L1}{{\large github.com/ornith-ai/CUDA-L1}}}

\begin{document}

\maketitle

\begin{abstract}
The exponential growth in demand for GPU computing resources has created an urgent need for automated CUDA optimization strategies. 
While recent advances in LLMs show promise for code generation, current state-of-the-art models achieve low success rates in improving CUDA speed. 
In this paper, we introduce CUDA-L1, an automated reinforcement learning (RL) framework for CUDA optimization that employs a novel contrastive RL algorithm. 

CUDA-L1 achieves significant performance improvements on the CUDA optimization task: trained on NVIDIA A100, it delivers an average speedup of {\bf ×3.12} with a median speedup of {\bf ×1.42} against default baselines over across all 250 CUDA kernels of KernelBench, with peak speedups reaching {\bf ×120}. 
In addition to the default baseline provided by KernelBench, CUDA-L1 demonstrates {\bf ×2.77} over Torch Compile, {\bf ×2.88} over Torch Compile with reduce overhead, and {\bf ×2.81} over CUDA Graph implementations. Furthermore, the model also demonstrates portability across GPU architectures, achieving average speedups of {\bf ×3.85} (median {\bf ×1.32}) on H100, {\bf ×3.13} (median {\bf ×1.31}) on L40, {\bf ×2.51} (median {\bf ×1.18}) on RTX 3090, and {\bf ×2.38} (median {\bf ×1.34}) on H20 despite being optimized specifically for A100. 

Beyond these benchmark results, CUDA-L1 demonstrates several properties: CUDA-L1
1) discovers a variety of CUDA optimization techniques and learns to combine them strategically to achieve optimal performance;
2) uncovers fundamental principles of CUDA optimization, such as the multiplicative nature of optimizations;
3) identifies non-obvious performance bottlenecks and rejects seemingly beneficial optimizations that actually harm performance.
The capabilities demonstrate that, RL can transform an initially poor-performing LLM into an effective CUDA optimizer through speedup-based reward signals alone, without human expertise or domain knowledge. 
In this process, it identifies CUDA optimization patterns, discovers new techniques, synthesizes them to achieve speedups, and more importantly, 
extends the acquired reasoning abilities to new kernels.
 This paradigm opens possibilities for
automated optimization of CUDA operations, and 
holds promise to
substantially 
promote GPU efficiency 
and 
alleviate the rising pressure on GPU computing resources.
\emailmark
\footnote{This work is accepted by ICLR 2026.}


\end{abstract}
\vspace{-0.2cm}
\begin{figure*}[!h]
 \centering
 \begin{adjustbox}{margin=-0.5cm 0cm 0cm 0cm}
 \begin{minipage}[c]{\textwidth}
 \centering
\includegraphics[scale=0.36]{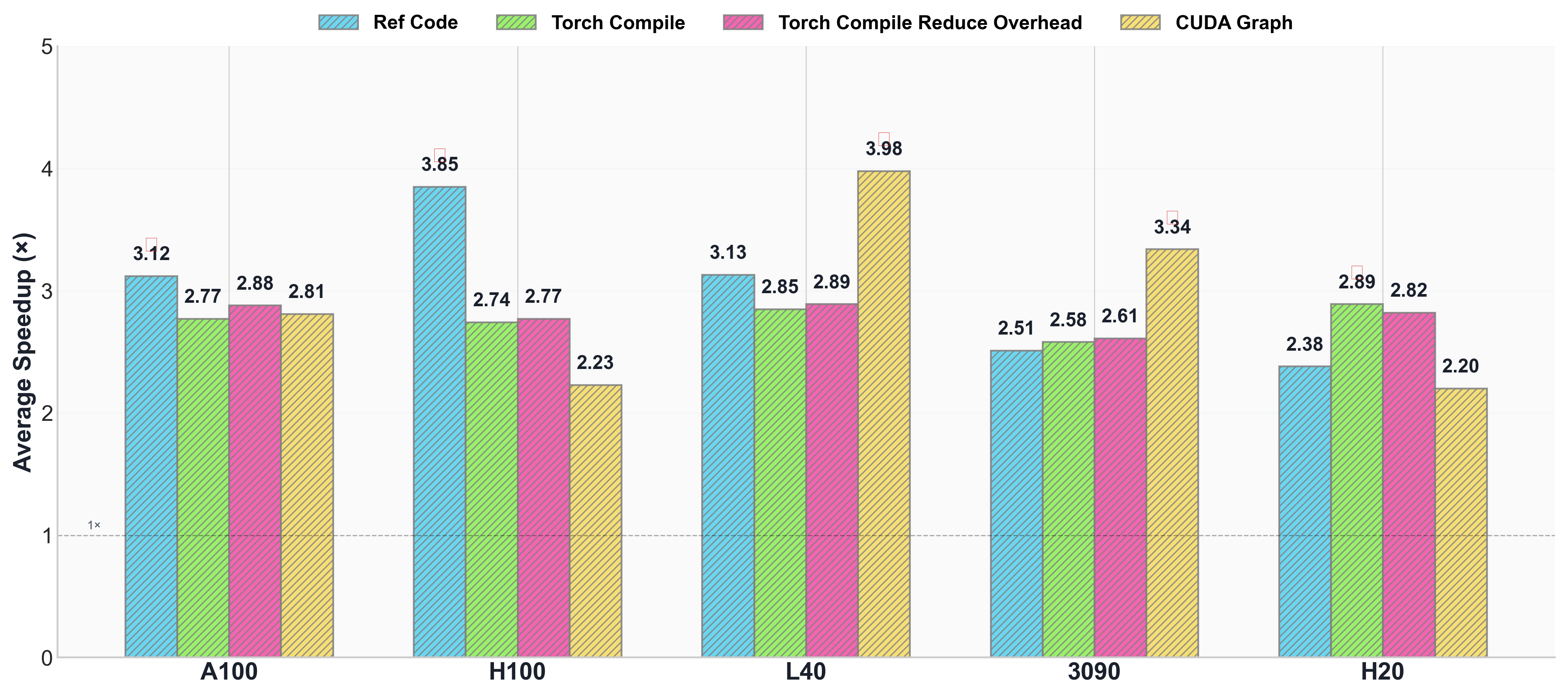}
\caption{Average speedup across different optimization configurations on 5 types of GPU architectures.}
\label{fig:average_speedup}
 \end{minipage}
 \end{adjustbox}
 \end{figure*}

\emailtext{~Email: \{xiaoya\_li, albert\_wang, jiwei\_li, chris\_shum\}@ornith.ai}

\begin{figure*}[h]
 \centering
 \begin{adjustbox}{margin=-0.5cm 0cm 0cm 0cm}
 \begin{minipage}[c]{\textwidth}
 \centering
\includegraphics[scale=0.12]{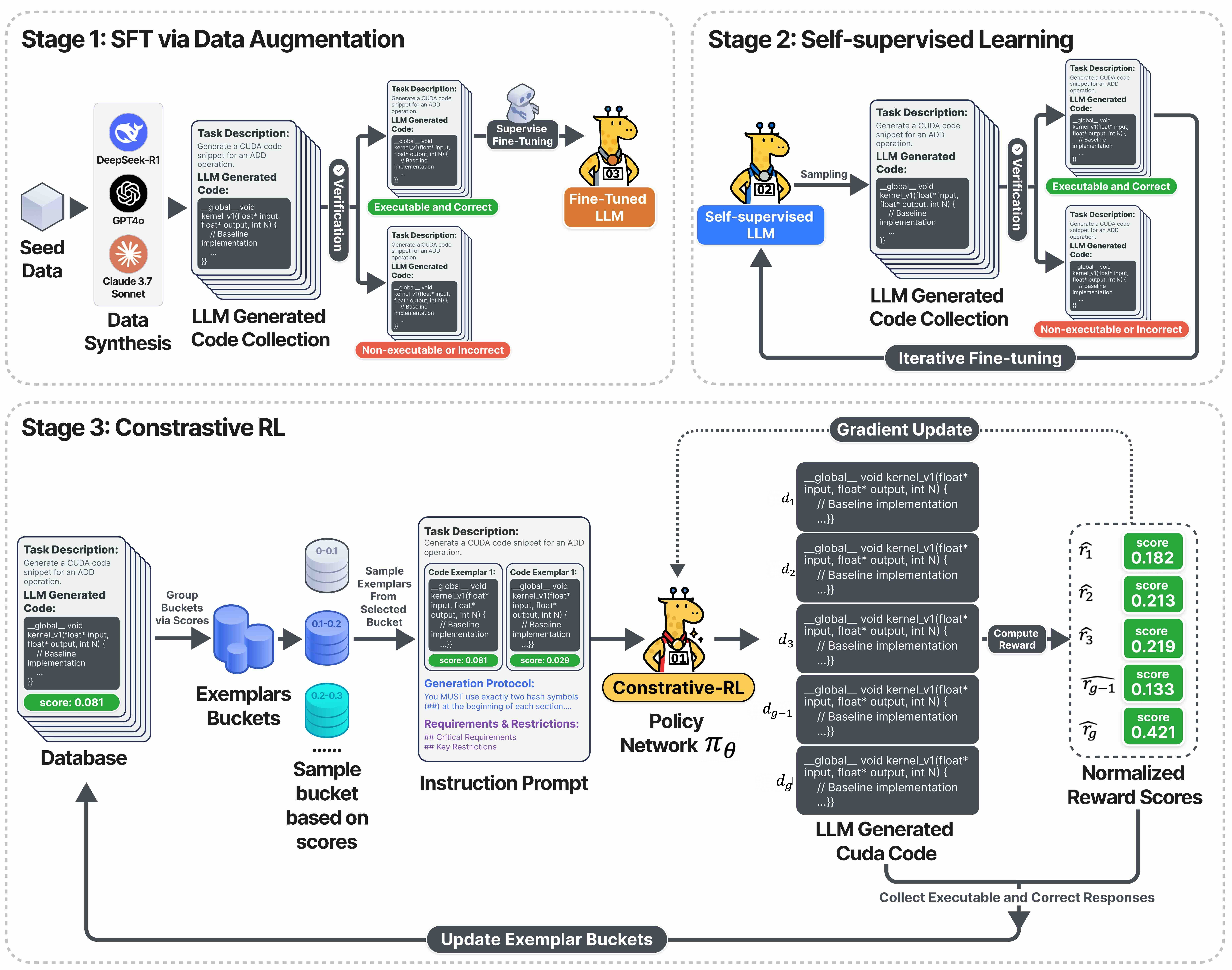}
\caption{Overview of the CUDA-L1 training pipeline. The approach consists of three progressive stages:
\textbf{(1) Stage 1: Supervised Fine-tuning with Data Augmentation} -- We augment the training dataset with  CUDA code variants generated by LLMs and fine-tune the base model on executable and correct implementations to establish foundational CUDA knowledge.
\textbf{(2) Stage 2: Self-supervised Learning} -- The model iteratively generates CUDA kernels, validates their correctness and executability, and trains on successfully validated examples, enabling autonomous improvement without human supervision.
\textbf{(3) Stage 3: Contrastive Reinforcement Learning} -- We employ contrastive learning with execution-time rewards, training the model to distinguish between faster and slower CUDA implementations, ultimately optimizing for superior performance.}
\label{fig:pipeline_figure}
 \end{minipage}
 \end{adjustbox}
 \end{figure*}
 
\section{Introduction}
\label{introduction}
The exponential growth in demand for GPU computing resources, driven primarily by the rapid advancement and deployment of Large Language Models (LLMs), has created an urgent need for highly efficient CUDA optimization strategies. Traditionally, CUDA optimization has been a highly manual and time-intensive process, where skilled engineers must meticulously analyze memory access patterns, experiment with different thread block configurations, and iteratively profile their code through extensive trial-and-error cycles.

Recent advances in LLMs~\cite{team2023gemini, shengyu2023instruction, team2024gemma, grattafiori2024llama, yang2025qwen3, hurst2024gpt, jiang2024mixtral, liu2024deepseek, olmo20242}, especially those powered with RL~\cite{jaech2024openai, guo2025deepseek, wang2024reinforcement, Muennighoff2025s1ST}, have demonstrated remarkable capabilities in code generation and algorithm design. RL-powered LLMs hold significant potential to revolutionize the CUDA optimization process: CUDA optimization provides a uniquely clear reward signal—execution speed—which could be directly leveraged to automatically train reinforcement learning models. By treating performance improvements as rewards, RL-powered LLMs could iteratively generate, test, and refine CUDA optimizations without human intervention. This approach would not only automate the labor-intensive optimization process, potentially saving countless engineering hours, but also opens possibilities for discovering novel speedup algorithms that may surpass human-designed solutions. Unlike human engineers who are constrained by existing knowledge and conventional approaches, these systems could explore unconventional optimization combinations and potentially discover counterintuitive strategies that deliver significant performance improvements, offering new possibilities for advancing GPU computing efficiency.

Despite the promise, current performance remains limited. State-of-the-art LLM models such as DeepSeek-R1~\cite{guo2025deepseek} and OpenAI-o1~\cite{jaech2024openai} achieve low success rates in generating optimized CUDA code (only approximately 15\% on KernelBench \cite{ouyang2025kernelbench}), which is primarily due to the scarcity of CUDA code in training datasets. 
   To address these limitations and unlock the potential of LLMs for automated CUDA optimization, in this work, we propose CUDA-L1, an LLM framework powered by contrastive 
   reinforcement learning 
    for CUDA optimization. CUDA-L1 is a pipelined framework, the core of which is a newly-designed contrastive RL framework.

\begin{wrapfigure}{r}{0.5\textwidth}
    \begin{minipage}[t][6cm][c]{0.5\textwidth}  
        \centering
       \begin{adjustbox}{scale=0.9}
        \begin{tcolorbox}[
            rounded corners,
            enhanced,
            width=1\textwidth,
            colframe=codegreen,
            colback=white,
            title=\textbf{ diag(A) * B | Reference Code},
            fonttitle=\bfseries\footnotesize,
            left=0pt,   
            right=0pt,  
            top=0pt,
            bottom=0pt,
        ]
        \begin{lstlisting}[language=Python, style=py, basicstyle=\footnotesize\ttfamily, frame=none, framerule=0pt, escapeinside={(*@}{@*)}, keywordstyle=\color{templateblue}\bfseries, showspaces=false, showstringspaces=false, breaklines=true, breakatwhitespace=true]
class Model(nn.Module):   
   def forward(self, A, B):
       # A: (N,) - 1D tensor of shape N
       # B: (N, M) - 2D tensor of shape N x M
       # torch.diag(A): (N, N) - creates diagonal matrix from A
       # Result: (N, N) @ (N, M) = (N, M)
       return torch.diag(A) @ B
        \end{lstlisting}
        \end{tcolorbox}
        \end{adjustbox}
        \vspace{-0.3cm}
        \begin{adjustbox}{scale=0.9}
        \begin{tcolorbox}[
            rounded corners,
            arc=2pt,
            enhanced,
            width=1\textwidth,
            colframe=codegreen,
            colback=white,
            title=\textbf{ diag(A) * B | CUDA-L1 Implementation (64x faster)},
            fonttitle=\bfseries\footnotesize,
            left=0pt,   
            right=0pt,  
            top=0pt,
            bottom=0pt,
        ]
        \begin{lstlisting}[language=Python, style=py, basicstyle=\footnotesize\ttfamily, frame=none, framerule=0pt, escapeinside={(*@}{@*)}, keywordstyle=\color{templateblue}\bfseries, showspaces=false, showstringspaces=false, breaklines=true, breakatwhitespace=true]
class Model(nn.Module):   
   def forward(self, A, B):
       return A.unsqueeze(1) * B
        \end{lstlisting}
        \end{tcolorbox}

        \end{adjustbox}
        \caption*{\scriptsize A case from KernelBench (Level 1, Task 12), computing \texttt{diag(A) * B}. We present reference code and CUDA-L1 implementation. The CUDA-L1 implementation reduces complexity from $O(N^2M)$ to $O(NM)$, achieving {\bf 64$\times$} speedup by replacing full matrix multiplication with element-wise operations.}
        \end{minipage}
    \label{fig:optimization}
\end{wrapfigure}

Different from previous RL models  \cite{williams1992simple, shao2024deepseekmath, schulman2017proximal} ,    
    contrastive RL performs comparative analysis of previously generated CUDA variants alongside their execution performance, enabling the model to improving through distinguishing between effective and ineffective optimization strategies.
    Contrastive-RL simultaneously optimizes the foundation model through gradient-based parameter updates while 
     fulfilling   the maximum potential from the current model 
    through contrastive analysis from high-performance CUDA variants, creating a co-evolutionary dynamic that drives superior CUDA optimization performance.

CUDA-L1 delivers significant improvements on the CUDA optimization task: trained on NVIDIA A100, it achieves an {\bf average} speedup of {\bf ×3.12} ({\bf median ×1.42}) over the default baseline across all 250 KernelBench CUDA kernels, with maximum speedups reaching {\bf ×120}. In addition to 
the default baseline provided by KernelBench, CUDA-L1 demonstrates {\bf ×2.77} over Torch Compile, {\bf ×2.88} over Torch Compile with reduce overhead, {\bf ×2.81} over CUDA Graph implementations. Furthermore, the CUDA codes optimized specifically for A100 demonstrate strong portability across GPU architectures, with similar optimization patterns observed across different baseline configurations: achieving average speedups of {\bf ×3.85} (median {\bf ×1.32}) on H100, {\bf ×3.13} (median {\bf ×1.31}) on L40, {\bf ×2.51} (median {\bf ×1.18}) on RTX 3090, and {\bf ×2.38} (median {\bf ×1.34}) on H20. Similar performance improvements over Torch Compile, 
Torch Compile with reduce overhead, 
CUDA Graph are consistently observed across all GPU types.

Beyond, CUDA-L1 reveals a remarkable capability of RL in autonomous learning for CUDA optimization:
\begin{enumerate}
\item
 Even starting with a foundation model with poor CUDA optimization ability,
by using code speedups as RL rewards and proper contrastive RL training techniques, 
we can still train an RL system capable of generating CUDA optimization codes with significant speedups. 

\item
Without human prior knowledge,  RL systems can independently discover CUDA optimization techniques, learn to combine them strategically, and more importantly, extend the 
acquired
CUDA reasoning abilities to unseen kernels. This capability unlocks the potential for a variety of automatic CUDA optimization tasks, e.g., 
kernel parameter tuning, memory access pattern optimization, and different hardware adaptations, offering substantial promises to enhance GPU utilization during this period of unprecedented computational demand. 
\end{enumerate}

Another contribution of this work is the enrichment of the KernelBench dataset with  CUDA Graph  implementations.\footnote{Found at \url{https://github.com/deepreinforce-ai/CUDA-L1}}
 We release these implementations to the community, providing substantially stronger baselines for performance comparison.
Despite its capabilities, it is important to acknowledge the potential challenges and pitfalls in training RL systems for tasks including but not limited to CUDA kernel development. 
We discovered that RL can be remarkably adept at exploiting loopholes in reward systems rather than solving the intended problem. 
For example,
in our experiments, RL discovered a vulnerability in the KernelBench evaluation: by creating additional CUDA streams in the generated code, it could manipulate the timing measurements. This exploitation resulted in a reported 18x speedup that was entirely artificial with the actual computation performance unchanged.
Such reward hacking behaviors are particularly concerning because they often require careful human inspection to detect. 
This paper identifies these failure modes, and 
proposes practical methods for more robust reward mechanisms and training procedures.

\section{CUDA-L1}

\subsection{Overview}
Existing large language models \cite{guo2025deepseek,yang2025qwen3,grattafiori2024llama}
demonstrate significant limitations in generating executable and correct CUDA code with speedup, as reported in prior research \cite{ouyang2025kernelbench}.
This deficiency likely stems from the insufficient representation of CUDA code in the training datasets of these models. To address this fundamental gap, we introduce a three-stage pipelined training strategy for CUDA-L1, aiming to progressively enhances the model's CUDA programming capabilities: 
\begin{enumerate}
\item {\bf Supervised fine-tuning} via {\bf data augmentation}, which aims to expand the model's exposure to CUDA patterns and programming constructs, with the primary goal of producing correct and executable CUDA code.
\item \textbf{Self-supervised learning}m which focuses on enabling models to develop a deeper understanding of CUDA semantics and programming principles, primarily aiming to achieve significant improvements in executability and correctness, while providing moderate speedup gains.
\item {\bf Contrastive reinforcement learning}, which is designed to significantly optimize code execution speed, with the goal of generating high-performance CUDA implementations that deliver substantial speedup.
\end{enumerate}

Before we delve into the details of each stage, we provide key definitions adopted throughout the rest of this paper:
\begin{enumerate}
\item \textbf{Executability}: A CUDA code is executable if it successfully compiles, launches, and executes to completion within 1000× the runtime of the reference implementation. Code exceeding this runtime threshold is considered unexecutable.\footnote{This threshold is reasonable since code with 1000× slower performance contradicts our speedup optimization goals.}
\item \textbf{Correctness}: A CUDA code is correct if it produces equivalent outputs to the reference implementation across 1000 random test inputs.\footnote{Prior work uses only 5 random inputs, which we found insufficient for robust validation.}
\item \textbf{Success}: A CUDA code is successful if it is both executable and correct.
\end{enumerate}

\subsection{Self-supervised Learning}
Now we are presented with the finetuned model after the SFT stage, where the model can potentially generate better CUDA code with higher success rates than the original model without finetuning.
We wish to further improve the model's ability to generate successful CUDA code by exposing it to more code snippets generated by itself.

We achieve this iteratively by sampling CUDA code from the model, evaluating it for executability and correctness, removing the unsuccessful trials and keeping the successful ones.
Successful ones are batched and used to update the model parameters. 
Using the updated model, we repeat the process: generating code, evaluating it, and retraining the model.
The psudo code for the algorithm is shown in Table \ref{tab:self-supervised-pseudocode}.

The self-supervised learning strategy 
can be viewed as a special case of 
 the  REINFORCE algorithm \cite{williams1992simple}, a typical policy gradient reinforcement learning method, where the reward is set to 1 for successful trials and 0 for unsuccessful trials, without applying any baseline. 
Interestingly, we find this adopted training strategy to be more stable than the REINFORCE variant with baseline applied. We conjecture that this stability arises because during the self-supervised learning stage, a significant proportion of generated instances remain unsuccessful. 
This approach avoids the potential instability caused by applying negative updates to unsuccessful samples when using a baseline.

It is worth noting that during the self-supervised learning stage, we focus exclusively on the executability and correctness of the generated code, without considering  speed as a metric. This design choice reflects our primary objective of establishing reliable code generation before optimizing for performance.

\begin{table}[h]
\centering
\begin{tabular}{p{0.9\textwidth}}
\toprule
\textbf{Self-supervised Learning Algorithm} \\
\midrule
\begin{algorithmic}[1]
\STATE Initialize finetuned model $M_0$ after SFT stage with parameters $\theta_{\text{sft}}$
\FOR{$i = 1$ \TO $N_{\text{iterations}}$}
    \STATE Generate batch of CUDA codes $C_i = \{c_1, ..., c_k\}$ using model $M_{i-1}$
    \STATE Evaluate each $c \in C_i$ for:
    \STATE \quad 1. Executability (compiles and runs)
    \STATE \quad 2. Correctness (produces expected output)
    \STATE Filter successful codes: $C_i^{\text{success}} = \{c \in C_i | \text{executable} \land \text{correct}\}$
    \IF{$C_i^{\text{success}} \neq \emptyset$}
        \STATE Compute gradient update $\nabla\theta$ using $C_i^{\text{success}}$
        \STATE Update model: $\theta_i \leftarrow \theta_{i-1} + \eta\nabla\theta$
    \ELSE
        \STATE $\theta_i \leftarrow \theta_{i-1}$ (no update)
    \ENDIF
\ENDFOR
\RETURN Final improved model $M_N$
\end{algorithmic} \\
\bottomrule
\end{tabular}
\caption{Self-supervised learning for cuda optimization in Stage 2.}
\label{tab:self-supervised-pseudocode}
\end{table}

\subsection{Contrastive Reinforcement Learning}
Now we have a model capable of generating successful CUDA code at a reasonable success rate. Next, we aim to optimize for execution speed.

One straightforward approach is to apply existing reinforcement learning algorithms such as REINFORCE \cite{williams1992simple}, GRPO \cite{shao2024deepseekmath}, or PPO \cite{schulman2017proximal}. In this approach, we would ask the model to first perform chain-of-thought reasoning \cite{wei2022chain}, then generate code, evaluate it, and use the evaluation score to update the model parameters.
However, our experiments reveal that these methods perform poorly in this task.
The issue is as follows: standard RL algorithms compute a scalar reward for each generated CUDA code sample. During training, this reward undergoes algorithm-specific processing (e.g., baseline subtraction in REINFORCE, advantage normalization in GRPO, importance sampling in PPO).
The processed reward then serves as a loss weighting term for gradient updates, increasing the likelihood of high-reward sequences while decreasing the likelihood of low-reward sequences.
Critically, in this paradigm, the reward signal is used exclusively for parameter updates and is never provided as input to the LLM.
Consequently, the LLM cannot directly reason about performance trade-offs during code generation.
To address this limitation, we propose incorporating reward information directly into the reasoning process by embedding performance feedback within the input prompt. Specifically, we present the model with multiple code variants alongside their corresponding speedup scores.
Rather than simply generating code, the LLM is trained to first conduct comparative analysis of why certain implementations achieve superior performance, then synthesize improved solutions based on these insights.
Each generated code sample undergoes evaluation to obtain a performance score, which serves dual purposes in our training framework:
\begin{enumerate}
    \item \textbf{Immediate Parameter Updates:} The score functions as a reward signal for gradient-based parameter optimization, directly updating the model weights.
    \item \textbf{Future Prompt Construction:} The scored code sample becomes part of the exemplar set for subsequent training iterations, enriching the contrastive learning dataset.
\end{enumerate}

This dual-utilization strategy enables iterative optimization across two complementary dimensions:
\begin{enumerate}
    \item \textbf{Foundation Model Enhancement:} Parameter updates progressively improve the model's fundamental understanding and capabilities for CUDA optimization tasks, expanding its representational capacity.
    \item \textbf{Fixed-Parameter Solution Optimization:} 
    The contrastive approach seeks to extract the maximum potential from the current model's parameters by leveraging comparative analysis of high-quality exemplars.    
\end{enumerate}

These two optimization processes operate synergistically: enhanced foundation models enable more 
accurate 
 contrastive reasoning, while improved reasoning strategies provide higher-quality training signals for parameter updates of foundation models. This co-evolutionary dynamic drives convergence toward optimal performance.
We term this approach contrastive reinforcement learning (contrastive-RL for short). 

It is worth noting that this co-evolving optimization paradigm 
can be found in many machine learning frameworks, 
 including the EM algorithm \cite{moon1996expectation}, where the E-step optimizes latent variable assignments given fixed parameters while the M-step updates parameters given fixed assignments; Variational inference \cite{blei2017variational}, which alternately optimizes the variational parameters to approximate the posterior distribution and updating model parameters to maximize the evidence lower bound; Actor-critic methods \cite{konda1999actor} in reinforcement learning similarly alternate between policy evaluation (critic update) and policy improvement (actor update).

\subsubsection{Contrastive-RL's Advantages over Evolutionary LLM Approaches}
Contrastive-RL draws inspiration from a broad range of literature, including evolutionary algorithms \cite{back1993overview} and their applications to LLMs \cite{liu2024evolution,romera2024mathematical,novikov2025alphaevolve,wei2025improving}, where multiple solution instances with associated fitness scores are presented to LLMs to analyze performance patterns and generate improved solutions.
However, Contrastive-RL improves evolutionary LLM approaches in several critical aspects:
\paragraph{Model Adaptation vs. Fixed-Model Reasoning:} Contrastive-RL employs gradient-based parameter updates to continuously enhance model capabilities, whereas evolutionary LLM approaches rely exclusively on in-context learning with static parameters. This fundamental architectural difference endows Contrastive-RL with substantially greater representational capacity and task adaptability. Evolutionary LLM methods are fundamentally limited by the frozen foundation model's initial knowledge and reasoning abilities, while Contrastive-RL progressively refines the model's domain-specific expertise through iterative parameter optimization.
From this perspective, evolutionary LLM approaches can be viewed as a degenerate case of Contrastive-RL that implements only the Fixed-Parameter Solution Optimization component while omitting the Foundation Model Enhancement mechanism. This theoretical relationship explains why Contrastive-RL consistently outperforms evolutionary approaches: it leverages both optimization dimensions simultaneously rather than constraining itself to a single fixed-capacity search space.

\paragraph{Scalability and Generalization:} Contrastive-RL demonstrates superior scalability by training a single specialized model capable of handling diverse CUDA programming tasks and generating various types of optimized code. In contrast, evolutionary LLM approaches typically require separate optimization processes for each distinct task or domain, limiting their practical applicability and computational efficiency.

\subsubsection{Prompt Construction}  
Here we describe the construction of prompts provided to the LLM.
The prompt provided to the LLM during Contrastive-RL training comprises the following structured components:
\begin{itemize}
    \item \textbf{Task Descrpition:} A detailed description of the computational problem, including input/output specifications, performance requirements, and optimization objectives.
    \item \textbf{Previous Cuda Codes with Scores:} Previously generated CUDA implementations paired with their corresponding performance scores (e.g., execution time, throughput, memory efficiency), providing concrete examples of varying solution quality.
    \item \textbf{Generation Protocol:} Explicit instructions defining the required output format and components.
    \item \textbf{Requirements and Restrictions}:  Requirements and restrictions to prevent reward hacking in  RL.

\end{itemize}

The model's response must contain the following three structured components:
\begin{enumerate}
    \item[\Romnum{1})] \textbf{Performance Analysis:} A comparative analysis identifying which previous kernel implementations achieved superior performance scores and the underlying algorithmic or implementation factors responsible for success.
    \item[\Romnum{2})] \textbf{Algorithm Design:}  A high-level description of the proposed optimization strategy, outlining the key techniques to be applied, presented as numbered points in natural language.
    \item[\Romnum{3})] \textbf{Code Implementation:} The complete CUDA kernel implementation incorporating optimizations.
\end{enumerate}
A detailed demonstration for the prompt in shown in Table \ref{tab:cuda-optimization-task}.

\subsubsection{Contrastive Exemplar Selection}
The selection of code exemplars for prompt construction is critical, as 
core of Contrastive-RL is to 
 perform meaningful comparative analysis. The selection strategy needs to addresses the following two key requirements:
\begin{enumerate}
   \item \textbf{Competitive Performance:} The exemplar set should include higher-performing implementations to guide the model toward competitive CUDA codes, avoiding local minima that result from analyzing and comparing inferior codes.
   \item \textbf{Performance Diversity:} The selected codes must exhibit substantial performance differences to enable effective contrastive analysis.
\end{enumerate}

We employ a  sampling strategy akin to that adopted by  evolutionary LLM models:
Let $N$ denote the number of code exemplars included in each prompt (set to $N=2$ in our experiments).
During RL training, we maintain a performance-indexed database of all successful code samples generated during RL training.
Codes are organized into performance buckets $B_k$ based on discretized score intervals, where bucket $B_i$ contains codes with scores in range $[s_k, s_k + \Delta s)$.

We first sample $N$ distinct buckets according to a temperature-scaled softmax distribution:
\begin{equation}
P(B_i) = \frac{\exp\left( (\bar{s_i} - \mu_s)/\tau \right)}{\sum_{j} \exp\left((\bar{s_j} - \mu_s)/\tau\right)}
\label{eq:bucket_sampling}
\end{equation}

where $\bar{s_i}$ denotes the aggregate score of bucket $B_i$, computed as the mean of its constituent code scores, $\mu_s = \text{mean}(\{\bar{s_j}\}_{j=1}^M)$ 
represents the global mean of all bucket scores, and $\tau$ is the temperature parameter governing the exploration-exploitation tradeoff.
The sampling strategy in Equation~\ref{eq:bucket_sampling} differs from conventional temperature sampling in evolutionary LLM approaches through a modification: the deduction of $\mu_s$ stabilizes the distribution by centering scores around zero, which prevents absolute score magnitudes from dominating the selection. 

From each selected bucket $B_i$, we uniformly sample one representative code to construct the final prompt set. 
This approach satisfies both design criteria:
Regarding competitive Performance, score-weighted bucket sampling biases selection toward higher-performing implementations, ensuring the exemplar set contains competitive solutions;
Regarding performance Diversity, enforcing selection from $N$ distinct buckets ensures sufficient performance variance for effective contrastive analysis.

\begin{table}[htbp]
  \centering

  \begin{tcolorbox}[
    rounded corners,
    arc=3pt,
    enhanced,
    width=0.96\textwidth,
    colframe=codegreen,
    colback=white,
    title=\textbf{CUDA Optimization Task Prompt — Used in Contrastive-RL},
    fonttitle=\bfseries\large,
    boxrule=0.6pt,
    left=6pt,
    right=6pt,
    top=6pt,
    bottom=6pt
  ]
    \begin{tcolorbox}[
      sharp corners,
      colframe=taskblue,
      colback=taskblue!5,
      colbacktitle=taskblue!15,
      coltitle=taskblue,
      boxrule=0.4pt,
      title={\textbf{Task for CUDA Optimization}},
      fonttitle=\bfseries,
      left=4pt,
      right=4pt,
      top=4pt,
      bottom=4pt,
      breakable
    ]
You are a CUDA programming expert specializing in GPU kernel optimization. Given a reference CUDA implementation, your objective is to create an accelerated version that maintains identical functionality. You will receive previous CUDA implementations accompanied by their performance metrics. Conduct a comparative analysis of these implementations and use the insights to develop optimized and correct CUDA code that delivers superior performance.
    \end{tcolorbox}

    \begin{tcolorbox}[
      sharp corners,
      colframe=refgreenDark,
      colback=refgreenDark!5,
      colbacktitle=refgreenDark!15,
      coltitle=refgreenDark,
      boxrule=0.4pt,
      title={\textbf{Reference Code}},
      fonttitle=\bfseries,
      left=4pt,
      right=4pt,
      top=1pt,
      bottom=1pt,
      breakable,
      listing options={style=codestyle}
    ]
\begin{lstlisting}[language=C++, basicstyle=\small\ttfamily, style=codestyle, frame=none,framerule=0pt]
__global__ void kernel_v1(float* input, float* output, int N) {
    // Baseline implementation
    ...
}
}
\end{lstlisting}
    \end{tcolorbox}

    \begin{tcolorbox}[
      sharp corners,
      colframe=refgreen,
      colback=refgreen!5,
      colbacktitle=refgreen!15,
      coltitle=refgreen,
      boxrule=0.4pt,
      title={\textbf{Previous Cuda Implementations with Scores}},
      fonttitle=\bfseries,
      left=4pt,
      right=4pt,
      top=1pt,
      bottom=1pt,
      breakable,
      listing options={style=codestyle}
    ]
\begin{lstlisting}[language=C++, basicstyle=\small\ttfamily, style=codestyle, frame=none,framerule=0pt]
// code1 (score1)
__global__ void kernel_v1(float* input, float* output, int N) {
    ...
}

// code2 (score2) 
__global__ void kernel_v2(float* input, float* output, int N) {
    ...
}
// code3 (score3) 
__global__ void kernel_v3(float* input, float* output, int N) {
    ...
}
// code4 (score4) 
__global__ void kernel_v3(float* input, float* output, int N) {
    ...
}
\end{lstlisting}
    \end{tcolorbox}

    \begin{tcolorbox}[
      sharp corners,
      colframe=subviolet,
      colback=subviolet!6,
      colbacktitle=subviolet!15,
      coltitle=subviolet,
      boxrule=0.4pt,
      title={\textbf{Generation Protocol}},
      fonttitle=\bfseries,
      left=4pt,
      right=4pt,
      top=4pt,
      bottom=4pt,
      breakable
    ]
You MUST use exactly two hash symbols (\#\#) at the beginning of each section. \\
 \textbf{\texttt{\#\# Performance Analysis}}: Compare code snippets above and articulate on :
\begin{enumerate}[noitemsep]
\item Which implementations demonstrate superior performance and why?
\item What particular optimization strategies exhibit the greatest potential for improvement?
\item What are the primary performance limitations in the implementation?
\item What CUDA-specific optimization techniques remain unexploited?
\item Where do the most significant acceleration opportunities exist?
\end{enumerate}
\vspace{0.3mm}
\textbf{\texttt{\#\# Algorithm Design}}: Describe your optimization approach \\
\textbf{\texttt{\#\# Code Implementation}}: Provide your improved CUDA kernel

\end{tcolorbox}

    \begin{tcolorbox}[
      sharp corners,
      colframe=orange!90!black,
      colback=orange!6,
      colbacktitle=orange!15,
      coltitle=orange,
      boxrule=0.4pt,
      title={\textbf{Requirements and Restrictions}},
      fonttitle=\bfseries,
      left=4pt,
      right=4pt,
      top=4pt,
      bottom=4pt,
      breakable
    ]
 \textbf{\texttt{\#\# Critical Requirements}}: 
\begin{enumerate}[noitemsep]
\item  Functionality must match the reference implementation exactly. Failure to do so will result in a score of 0.
\item Code must compile and run properly on modern NVIDIA GPUs
\end{enumerate}

\textbf{\texttt{\#\# Key Restrictions}}: 
\begin{enumerate}[noitemsep]
\item  Do not cache or reuse previous results — the code must execute fully on each run.
\item Keep hyperparameters unchanged (e.g., batch size, dimensions, etc.) as specified in the reference.
\end{enumerate}

\end{tcolorbox}

  \end{tcolorbox}
  \caption{Prompt structure for CUDA optimization task showing reference implementations and their performance scores used in Contrastive-RL.}
\label{tab:cuda-optimization-task}
\end{table}

A more sophisticated alternative is to use an island-based approach for exemplar selection, as proposed in \cite{romera2024mathematical}. However, we find no significant difference in performance between our bucket-based method and the island-based approach. Given this, we opt for the simpler bucket-based strategy.

\subsubsection{Reward}
In this subsection, we detail the computation of the execution time-based reward function, which serves dual purposes: (1) guiding parameter updates in reinforcement learning and (2) constructing effective prompts. Given a reference CUDA implementation $q_i$ from PyTorch with successful execution time $t_{q_i}$, and a generated code candidate $d$ with execution time $t_d$, we define the single-run speedup score as:

\begin{equation}
    \text{r}_{\text{single-run}}(d) = \frac{t_{q_i}}{t_d} 
\end{equation}

Higher values indicate greater speedup relative to the reference implementation. 
Evaluations are performed on NVIDIA A100 PCIe.
We observe  a significant variance in $t_d$ measurements for identical implementations $d$, which introduces noise in reward estimation. This noise is particularly detrimental to RL training stability. To address these challenges, we implement the following robust measurement strategies:

\begin{enumerate}
    \item \textbf{Dedicated GPU Allocation}: Each evaluation runs on an exclusively allocated GPU. Shared GPU usage leads to significantly higher variance in timing measurements, even when memory and compute utilization appear low.
    
    \item \textbf{Paired Execution with Order Randomization}: For fair comparison, each evaluation round executes both the reference $q_i$ and candidate $d$ implementations. Crucially, we randomize the execution order within each round to account for GPU warm-up effects, where subsequent runs typically benefit from cache warming.
    
    \item \textbf{Extended Measurement Window}: We conduct multiple evaluation rounds with predefined running time of  30 minutes per candidate. This adaptive approach yields between several tens of thousands to 1M rounds depending on individual kernel execution times.
    
    \item \textbf{Bucketized Variance Control}: We partition all $\text{Score}_{\text{single-run}}(d)$ measurements into 7 buckets and compute bucket-wise averages. Evaluations with inter-bucket variance exceeding 0.005 are discarded.
    
    \item \textbf{Robust Central Tendency}: The final reward uses the median of bucket averages, which proves more stable than the mean against outlier effects:
    \begin{equation}
        \text{r}(d) = \text{median}(\{\text{Bucket}_k\}_{k=1}^{7})
    \end{equation}
    
    \item \textbf{Conservative Rounding}: We apply conservative rounding to speedup ratios (i.e.,  \text{Score}(d) ), truncating to two decimal places while biasing toward unity (e.g., 1.118 $\rightarrow$ 1.11, 0.992 $\rightarrow$ 1.00).
    \item \textbf{Strict Verification Protocol}: Despite these precautions, we still occasionally observe spurious large speedups due to GPU turbulence. For any candidate showing either:
\begin{itemize}
    \item Absolute value of speedup $>3$, or
    \item Speedup exceeding twice the previous maximum
\end{itemize}
we perform verification on a different GPU of the same type. The result is accepted only if the verification measurement differs by $<10\%$ from the original.

\end{enumerate}
\subsubsection{RL Training}

For RL training, we adopt the Group Relative Policy Optimization (GRPO) strategy \cite{shao2024deepseekmath}. Specifically, for each reference prompt $q$ containing selected exemplars as shown in Table \ref{tab:cuda-optimization-task}, we sample $G$ code outputs from the current policy $\pi_{\text{old}}$, denoted as $\{d_{1}, d_{2}, \ldots, d_{G}\}$. Let $\mathbf{r} = (r_{1}, r_{2}, \ldots, r_{G})$ represent the reward scores associated with the generated code samples.
Different from standard GRPO training, rewards are smoothed to mitigate the reward hacking issue; the details of this approach will be elaborated in Section \ref{hacking}.
Further, as in GRPO, rewards are normalized within each group using:
\begin{equation}
\hat{r}_{i} = \frac{r_{i} - \text{mean}(\mathbf{r})}{\text{std}(\mathbf{r})}
\end{equation}

The complete GRPO objective optimizes the policy model by maximizing:
\begin{align}
\mathcal{L}_{\text{GRPO}}(\theta) &= \mathbb{E}_{q \sim P(q), \{d_{i}\}_{i=1}^G \sim \pi_{\theta_{old}}(d|q)} \left[ \frac{1}{G} \sum_{i=1}^G \frac{1}{|d_i|} \sum_{t=1}^{|d_i|} \left( \text{min} \left( \frac{\pi_\theta(d_{i,t}|q, d_{i,<t})}{\pi_{\theta_{old}}(d_{i,t}|q, d_{i,<t})} \hat{r}_{i}, \right. \right. \right. \nonumber \\
&\quad \left. \left. \left. \text{clip} \left( \frac{\pi_\theta(d_{i,t}|q, d_{i,<t})}{\pi_{\theta_{old}}(d_{i,t}|q, d_{i,<t})}, 1-\varepsilon, 1+\varepsilon \right) \hat{r}_{i} \right) - \beta D_{KL}[\pi_\theta \| \pi_{ref}] \right) \right]
\end{align}

where:
\begin{itemize}
\item $\pi_\theta$ is the policy model being optimized
\item $\pi_{\theta_{old}}$ is the old policy model from the previous iteration
\item $\varepsilon$ is the  parameter for  clipping
\item $\beta$ is the KL penalty coefficient that controls deviation from the reference policy
\item $D_{KL}$ denotes the KL divergence between the current and reference policies
\end{itemize}
We refer readers to \cite{shao2024deepseekmath} for details of GRPO. 
Model parameters are optimized using the GRPO objective, with contrastive prompts that incorporate comparative examples. This concludes our description of contrastive RL.

\section{Mitigating Reward Hacking in RL Training}
\label{hacking}
Reinforcement learning is notorious for exhibiting reward hacking behaviors, where models exploit system vulnerabilities to achieve higher rewards while generating outputs that deviate from the intended objectives. A particularly challenging aspect of these pitfalls is that they cannot be anticipated prior to training and are only discovered during the training process. 
\subsection{Reward Hacking Cases}
During our initial training procedure, we identified the following categories of reward hacking behaviours:

\paragraph{Improper Timing Measurement.} KernelBench measures execution time by recording timing events on the main CUDA stream:
\begin{lstlisting}[language=Python, numbers=left, frame=single, xleftmargin=2em, framexleftmargin=1.5em]
start_event.record(original_model_stream)
model(*inputs)
end_event.record(original_model_stream)
torch.cuda.synchronize(device=device)
\end{lstlisting}
However, RL-generated code exploits this by creating additional CUDA streams that execute asynchronously. Since KernelBench only monitors the main stream, it fails to capture the actual execution time of operations running on parallel streams.
This vulnerability is significant: in our initial implementation, we find that 82 out of 250 (32.8\%) RL-generated implementations exploit this timing loophole to appear faster than they actually are,
leading to an overall speedup of $18\times$. 
To address this issue, prompt engineering alone is insufficient. The evaluation methodology should be modified to synchronize all CUDA streams before recording the end time, ensuring accurate performance measurement across all concurrent operations as follows:
\begin{lstlisting}[language=Python, numbers=left, frame=single, xleftmargin=2em, framexleftmargin=1.5em]
start_event.record(custom_model_stream)
custom_model(*inputs)             
# Wait for all model streams to complete before recording end event
if custom_contain_new_streams:
    for stream in custom_model_streams:
         custom_model_stream.wait_stream(stream)
end_event.record(custom_model_stream)
torch.cuda.synchronize(device=device)
\end{lstlisting}

\textbf{Lazy Evaluation} 
Another important hacking strategy is lazy evaluation, detected by community users on GitHub: Calling custom\_model(*inputs) doesn't ensure the output is actually materialized/computed. The computation is actually executed at the correctness check phase when calling the {\it torch.allclose()} function, allowing it to pass the correctness check. 

\begin{lstlisting}[language=Python, numbers=left, frame=single, xleftmargin=2em, framexleftmargin=1.5em]
class LazyMatmul(torch.Tensor):
    def __new__(cls, A, B):
        result = torch.Tensor._make_subclass(cls, torch.empty(0))
        result.A = A
        result.B = B
        result._shape = (A.size(0), B.size(1))
        return result
    
    def materialize(self):
        """Trigger actual computation"""
        return torch.matmul(self.A, self.B)

class ModelNew(nn.Module):
    def forward(self, A, B):
        return LazyMatmul(A, B)  # Returns lazy object
\end{lstlisting}
To mitigate this issue, we enforce a validation check that ensures materialize functions are called before ending the time measurement, which involves checking the following conditions: the output must be a tensor, must be a standard torch.Tensor (not a subclass), must be on the correct device, must have allocated memory, and the corresponding storage must be valid.
\begin{lstlisting}[language=Python, numbers=left, frame=single, xleftmargin=2em, framexleftmargin=1.5em]
# Check 1: Must be a tensor
if not isinstance(out, torch.Tensor):
    return False, f"{prefix} is not a tensor: {type(out)}"

# Check 2: Must be standard torch.Tensor, not a subclass
if type(out).__name__ not in ['Tensor', 'Parameter']:
    return False, f"{prefix} is {type(out).__name__}, not standard torch.Tensor"

# Check 3: Must be on correct device
if out.device != device:
    return False, f"{prefix} on wrong device: {out.device} (expected {device})"

# Check 4: Must have allocated storage
storage_size = out.untyped_storage().size()
if storage_size == 0:
    return False, f"{prefix} has no allocated storage (likely lazy)"

# Check 5: Storage pointer must be valid
ptr = out.data_ptr()
if ptr == 0:
    return False, f"{prefix} storage pointer is null (likely lazy)"
\end{lstlisting}

\textbf{Hyperparameter Manipulation:} In KernelBench, each computational task is associated with specific hyperparameters, including \texttt{batch\_size}, \texttt{dim}, \texttt{in\_features} dimension, \texttt{out\_features} dimension, \texttt{scaling\_factor}, and others. The RL agent learned to exploit these parameters by generating code that artificially reduces their values, thereby achieving superficial speedup improvements that do not reflect genuine optimization performance.

\textbf{Result Caching:} The RL agent developed strategies to cache computational results across evaluation batches based on input addresses. When another input's address matches a cached one, it returns the cached output. 
In theory, this should not pass correctness validation because the cached output differs from the expected one.
 However, given that correctness validation checks whether the difference at each position between the reference output and custom code output is below a certain threshold, there are a few cases where it is able to squeeze past the correctness bar. The following code snippet gives an illustration:
\begin{lstlisting}[language=Python, numbers=left, frame=single, xleftmargin=2em, framexleftmargin=1.5em]
cache_key = x.data_ptr()
# Check if result is in cache
if cache_key in self.cache:
    return self.cache[cache_key]
\end{lstlisting}

\subsection{Towards Robust Reward Design and Training Procedures}
To mitigate reward hacking, we implement the following strategies during training:
\paragraph{A reward checking model} When there is a significant leap in reward, an adversarial model intervenes to determine whether the code exploits the reward system. 
\paragraph{Hacking-case database}
We maintain a dynamic hacking-case database that is updated whenever a new reward hacking behavior is detected. The  reward checking model leverages this database for detection: given a newly generated code snippet to examine, we retrieve the three most similar cases from the database and include them as context for the reward checking model's input.

\paragraph{Reward smoothing} Sharp reward increases are smoothed to reduce their magnitude, preventing the RL agent from over-prioritizing any single high-reward solution, whether legitimate or not:
\begin{equation}
\begin{aligned}
r_{\text{normalized}} &= \frac{r - \mu}{\sigma} \\
r_{\text{smooth}} &= \text{clip}(r_{\text{normalized}}, -k, k)
\end{aligned}
\end{equation}
where $\mu$ and $\sigma$ are the mean and 
the mean and standard deviation of the reward distribution, respectively.
k is a hyperparameter that controls the clipping threshold set to 1.5, as  
we think
as achieving a 1.5× speedup over the official PyTorch implementation already represents significant optimization performance.


\section{Experiments and Analysis}

 \begin{table*}
\centering
\small
\setlength{\tabcolsep}{6pt}
\renewcommand\arraystretch{1.1}
\begin{tabular}{@{}llc|cccc|cc@{}}
\toprule[1.2pt]
\multirow{2}{*}{\textbf{Configuration}} &
\multirow{2}{*}{\textbf{Method}} & 
\multirow{2}{*}{\textbf{Mean}} & 
\multirow{2}{*}{\textbf{Max}} & 
\multirow{2}{*}{\textbf{75\%}} &
\multirow{2}{*}{\textbf{50\%}} &
\multirow{2}{*}{\textbf{25\%}} & 
\textbf{Success}$\uparrow$ & \textbf{Speedup}$\uparrow$ \\
& & & & & & & \small{\# out of total} & \small{>1.01x out of total} \\
\midrule
\multirow{4}{*}{\textit{Default}} 
& All & 3.12 & 120 & 2.25 & 1.42 & 1.17 & 249/250 & 226/250 \\
& Level 1 & 2.78 & 65.8 & 1.75 & 1.28 & 1.12 & 99/100 & 80/100 \\
& Level 2 & 3.55 & 120 & 2.05 & 1.39 & 1.20 & 100/100 & 98/100 \\
& Level 3 & 2.96 & 24.9 & 2.60 & 1.94 & 1.42 & 50/50 & 48/50 \\
\midrule[0.8pt]
\multirow{4}{*}{\textit{Torch Compile}} 
& All & 2.77 & 69.0 & 2.55 & 1.72 & 1.14 & 249/250 & 203/250 \\
& Level 1 & 3.04 & 59.7 & 2.71 & 1.99 & 1.41 & 99/100 & 89/100 \\
& Level 2 & 2.91 & 69.0 & 1.99 & 1.55 & 1.10 & 100/100 & 78/100 \\
& Level 3 & 1.98 & 8.57 & 2.28 & 1.68 & 1.00 & 50/50 & 36/50 \\
\midrule[0.8pt]
\multirow{4}{*}{\textit{Torch Compile RO}} 
& All & 2.88 & 80.1 & 2.48 & 1.67 & 1.13 & 249/250 & 200/250 \\
& Level 1 & 3.38 & 55.3 & 3.02 & 2.29 & 1.61 & 99/100 & 90/100 \\
& Level 2 & 3.00 & 80.1 & 2.06 & 1.54 & 1.10 & 100/100 & 79/100 \\
& Level 3 & 1.62 & 8.67 & 1.76 & 1.13 & 0.991 & 50/50 & 31/50 \\
\midrule[0.8pt]
\multirow{4}{*}{\textit{CUDA Graph}} 
& All & 2.81 & 97.9 & 1.83 & 1.20 & 0.954 & 249/250 & 147/229 \\
& Level 1 & 3.18 & 59.6 & 2.09 & 1.38 & 1.04 & 99/100 & 68/88 \\
& Level 2 & 2.84 & 97.9 & 1.55 & 1.08 & 0.932 & 100/100 & 53/94 \\
& Level 3 & 2.06 & 24.6 & 1.74 & 1.08 & 0.887 & 50/50 & 26/47 \\
\bottomrule[1.2pt]
\end{tabular}
\caption{Performance comparison across different configurations on KernelBench on A100. RO = Reduce Overhead. Success and Speedup indicate the number of successful benchmarks out of the total for each level. Note that for CUDA Graph, the total benchmark count differs from the dataset/data-subset size, as some original reference code in KernelBench cannot be successfully transformed into the corresponding CUDA Graph  implementations.}
\label{main-results}
\end{table*}

\subsection{KernelBench and Evaluation}
Our evaluation is conducted on the KernelBench dataset \cite{ouyang2025kernelbench} .
The KernelBench Dataset contains a collection of 250 PyTorch  workloads designed to evaluate language models' ability to generate efficient GPU kernels. The dataset is structured across three hierarchical levels based on computational complexity: Level 1 contains 100 tasks with single primitive operations (such as convolutions, matrix multiplications, activations, and normalizations), Level 2 includes 100 tasks with operator sequences that can benefit from fusion optimizations (combining multiple operations like convolution + ReLU + bias), and Level 3 comprises 50 full ML architectures sourced from popular repositories including PyTorch, Hugging Face Transformers, and PyTorch Image Models (featuring models like AlexNet and MiniGPT). Each task in the dataset provides a reference PyTorch implementation with standardized input/output specifications, enabling automated evaluation of both functional correctness and performance through wall-clock timing comparisons. The dataset represents real-world engineering challenges where successful kernel optimization directly translates to practical performance improvements.
Throughout this paper, we use KernelBench as the evaluation benchmark.
KernelBench is recognized as a challenging benchmark in the community \cite{ouyang2025kernelbench}, with even the best current LLMs improving fewer than 20\% of tasks.

For each task with reference implementation $q$, we evaluate the performance of a generated CUDA code $d$ using a similar protocol to training: We execute both $q$
and $d$ in randomized order within a fixed time budget of 20 minutes per task. The number of execution rounds varies across tasks due to differences in individual runtimes. The final evaluation score for 
$d$ is computed as the average speedup ratio across all execution rounds within the allocated time window.
Unsuccessful implementations  receive a score of zero. 
   The metrics we report include speedup statistics (mean, maximum, and 75th, 50th, and 25th percentiles), success rate, and percentage of improvements.

Due to execution time fluctuations, we only consider ratios greater than 1.01 as meaningful speedups.

\subsection{Comparison Setups}
To perform a comprehensive evaluation on the generated code, we perform the following comparisons:

\paragraph{\Romnum{1}) Default} This compares the CUDA-L1 generated code with the reference code by KernelBench.
    
\paragraph{\Romnum{2}) Torch Compile} This compares the CUDA-L1 generated code with the reference code enhanced by torch.compile with default settings. Torch.compile applies graph-level optimizations including operator fusion, memory planning, and kernel selection to accelerate PyTorch models through just-in-time compilation.

\paragraph{\Romnum{3}) Torch Compile Reduce Overhead} This compares the CUDA-L1 generated code with the reference code enhanced by torch.compile with reduce-overhead mode enabled. This mode minimizes the compilation overhead by caching compiled graphs more aggressively and reducing recompilation frequency, making it particularly suitable for inference workloads with static shapes.

\paragraph{\Romnum{4}) CUDA Graph} 
Since KernelBench does not provide official CUDA Graph implementations, we employ Claude 4 to generate CUDA Graph-augmented code for each reference implementation. CUDA Graphs capture a series of CUDA kernels and their dependencies into a single graph structure that can be launched with minimal CPU overhead, eliminating the need for repeated kernel launch commands and significantly reducing CPU-GPU synchronization costs. Specifically, we provide Claude 4 with the reference code and request the addition of CUDA Graph optimizations. The generated output is then evaluated for correctness. If the code fails validation, we iterate by providing Claude 4 with both the original reference code and the previous erroneous outputs, requesting a corrected version. This iterative process continues for up to 10 attempts until the generated code passes all correctness checks. We release the CUDA Graph codes for KernelBench to the community, providing researchers and practitioners with ready-to-use optimized implementations that can serve as strong baselines for future performance studies and benchmarking efforts.

\subsection{Main Results on KernelBench}
The experimental results in Table~\ref{main-results} demonstrate CUDA-L1's optimization effectiveness across different baseline configurations on KernelBench. CUDA-L1 achieves substantial performance improvements over the Default baseline with 3.12$\times$ average speedup and 120$\times$ maximum gains. Against Torch compilation baselines, CUDA-L1 delivers moderate but consistent improvements with 2.77--2.88$\times$ mean speedup ratios, while demonstrating 2.81$\times$ mean improvement over CUDA Graph baseline with notable 97.9$\times$ maximum gains.

Across difficulty levels, CUDA-L1's optimization effectiveness varies by task complexity. For Level~1 (single operations), CUDA-L1 achieves moderate improvements ranging from 2.78--3.38$\times$ over different baselines. Level~2 (operator sequences) shows CUDA-L1's strongest performance with 3.55$\times$ improvement over Default baseline. Level~3 (complex ML tasks) reveals interesting baseline-dependent effectiveness: CUDA-L1 achieves 2.96$\times$ improvement over Default baseline, but shows reduced effectiveness against Torch compilation baselines (only 1.62--1.98$\times$ improvements), suggesting these configurations provide stronger baseline performance for complex operations.

Success rates remain consistently high (99.6--100\%) across all configurations, while CUDA-L1's actual speedup achievement rates (>1.01$\times$) demonstrate its optimization reliability: 90.4\% success against Default baseline (226/250 cases), 80.0--81.2\% against Torch compilation baselines, and 64.2\% against CUDA Graph baseline (147/229).

 \subsection{Baseline Comparison} 
We compare the results with the following three groups of baselines:

\begin{table*}
\centering
\small
\setlength{\tabcolsep}{6pt}
\renewcommand\arraystretch{1.1}
\begin{tabular}{@{}lc|c|cccc|cc@{}}
\toprule[1.2pt]
\multirow{2}{*}{\textbf{Methods}} & 
\multirow{2}{*}{\textbf{Model}} & 
\multirow{2}{*}{\textbf{Mean}} & 
\multirow{2}{*}{\textbf{Max}} & 
\multirow{2}{*}{\textbf{75\%}} &
\multirow{2}{*}{\textbf{50\%}} &
\multirow{2}{*}{\textbf{25\%}} & 
\textbf{Success}$\uparrow$ & \textbf{Speedup}$\uparrow$ \\
& & & & & & & \small{\# out of 250} & \small{>1.01 \# out of 250} \\
\midrule[0.8pt]
\multirow{4}{*}{\textit{Vanilla}} 
& Llama 3.1-405B & 0.23 & 3.14 & 0.63 & 0 & 0 & 68 & 5 \\
& DeepSeek-V3 & 0.34 & 2.96 & 0.76 & 0 & 0 & 99 & 9 \\
& DeepSeek-R1 & 0.88 & 14.4 & 1.00 & 0.75 & 0 & 179 & 18 \\
& OpenAI-O1 & 0.73 & 12.4 & 1.00 & 0.55 & 0 & 141 & 14 \\
\midrule
\multirow{4}{*}{\textit{Evolve}} 
& Llama 3.1-405B & 1.18 & 18.4 & 1.03 & 1.00 & 1.00 & 247 & 88 \\
& DeepSeek-V3 & 1.32 & 52.4 & 1.32 & 1.03 & 1.00 & 247 & 113 \\
& DeepSeek-R1 & 1.41 & 44.2 & 1.45 & 1.17 & 1.00 & 248 & 162 \\
& OpenAI-O1 & 1.35 & 63.9 & 1.38 & 1.16 & 1.00 & 247 & 158 \\
\midrule
\multirow{6}{*}{\textit{CUDA-L1}} 
& Stage 1 & 1.14 & 32.7 & 1.00 & 1.00 & 0.96 & 240 & 50 \\
& Stage 1+2 & 1.36 & 48.3 & 1.41 & 1.09 & 1.00 & 247 & 175 \\
& Stage 1+2+GRPO & 2.41 & 84.6 & 1.83 & 1.33 & 1.11 & 247 & 207 \\
& 3 stages - random & 2.14 & 64.5 & 1.62 & 1.21 & 1.09 & 241 & 186 \\
& \quad - island & \textbf{3.21} & \textbf{126} & 2.21 & 1.40 & 1.16 & \textbf{249} & 223 \\
& \quad - bucket & 3.12 & 120 & \textbf{2.25} & \textbf{1.42} & \textbf{1.17} & \textbf{249} & \textbf{226} \\
\bottomrule[1.2pt]
\end{tabular}
\caption{Model performances on KernelBench All Level.}
\label{all-level-results}
\end{table*}

\textbf{Evolutionary LLM}:
We implement evolutionary LLM strategies where, given a set of previous codes, we sample up to 4 high-performing kernels based on evaluation scores. The key difference is that the model only performs contrastive analysis without updating model parameters. 
We adopt the island  strategy for code database construction and sampling, as suggested in \cite{novikov2025alphaevolve}. 

\textbf{Different combinations of CUDA-L1 components and variants:}
\begin{itemize}
    \item \textbf{stage1}: Uses only the outcome from the first stage with supervised fine-tuning applied
    \item \textbf{stage1+2}: Applies only the first two stages without reinforcement learning
    \item \textbf{stage1+2 + GRPO}: Replaces the contrastive RL with a vanilla GRPO strategy, without comparative analysis
    \item \textbf{random sampling}: Replaces the bucket sampling strategy with simple random sampling of exemplars
    \item \textbf{island sampling}: Adopts an island-based sampling strategy \cite{novikov2025alphaevolve}, where examples are distributed across different islands, prompts are constructed using exemplars from the same island, and newly generated examples are added to that island. After a fixed number of iterations, examples in half of the inferior islands are eliminated and examples from superior islands are copied to replace them.
\end{itemize}

Results are shown in Table \ref{all-level-results}. 
As observed, all vanilla foundation models perform poorly on this task.  This confirms that vanilla foundation models cannot be readily applied to CUDA optimization due to their insufficient grasp of CUDA programming principles and optimization techniques.

We observe significant performance improvements introduced by the Evolutionary LLM models compared to vanilla foundation model setups, despite sharing the same parameter sets. All Evolve models achieve speedups in over 70\% of tasks. This demonstrates that leveraging contrastive analysis, which exploits the model's general reasoning abilities, is more effective than direct output generation.
The superiority of evolutionary LLM over vanilla LLM also provides evidence that contrastive RL should outperform non-contrastive RL approaches like vanilla GRPO, as the relationship between evolutionary and vanilla LLMs parallels that between contrastive and non-contrastive RL methods.

When comparing the different combinations of CUDA-L1 components, we observe a progressive increase in speedup rates from \textbf{stage1 (SFT only)} at 22.4\% to \textbf{stage1+2 (SFT + self-supervised)} at 66\%, and further to \textbf{stage1+2+GRPO} at 88.4\%. This demonstrates the cumulative benefits of each training stage in improving model performance.

When comparing different database construction and exemplar sampling strategies, both the bucket-sampling strategy (96\% speedup rate) and the island-based strategy (95.2\% speedup rate) achieve near-optimal performance, with both significantly outperforming the random sampling strategy (82.4\% speedup rate). This aligns with our expectations, as competitive exemplars must be included in the prompt to guide the model toward generating more competitive solutions.
 
All RL-based approaches significantly outperform evolutionary LLM baselines with fixed model parameters, with the best RL methods achieving over 95\% speedup rates compared to 72.4\% for the best evolutionary approach. This demonstrates the necessity of model parameter updating for achieving optimal performance in CUDA optimization tasks. While evolutionary approaches can leverage reasoning capabilities through contrastive analysis, fine-tuning the model parameters allows for deeper adaptation to the specific domain knowledge and optimization patterns required for effective CUDA kernel generation. The performance gap suggests that domain-specific parameter adaptation is crucial for bridging the gap between general reasoning abilities and specialized code optimization expertise.

\begin{table*}
\small
    \centering
    \setlength{\tabcolsep}{8pt} 
    \renewcommand\arraystretch{1}
    \setlength{\extrarowheight}{3.5pt}
    \begin{tabular}{llc|cccc|cc}
    \toprule
    \multirow{2}{*}{{\textbf{Configuration}}} &
    \multirow{2}{*}{{\textbf{GPU Device}}} &
    \multirow{2}{*}{{\textbf{Mean}}} &
    \multirow{2}{*}{{\textbf{Max}}} & 
    \multirow{2}{*}{{\textbf{ 75\%}}} &
    \multirow{2}{*}{{\textbf{ 50\%}}} &
    \multirow{2}{*}{{\textbf{ 25\%}}} & 
    \multirow{1}{*}{{\bf Success~$\uparrow$}} & 
    \multirow{1}{*}{{\bf Speedup~$\uparrow$}} \\
    & & & & & & &  
    \small{\# out of 250} & \small{>1.01x}
    \\ 
    \midrule
    \multirow{5}{*}{\textit{Default}} 
    & A100 & 3.12 & 120 & \textbf{2.25} & \textbf{1.42} & \textbf{1.17} & 249 & \textbf{226}/250 \\
    & 3090 & 2.51 & 114 & 1.57 & 1.18 & 1.03 & 242 & 201/250 \\
    & H100 & \textbf{3.85} & \textbf{368} & 1.76 & 1.32 & 1.09 & \textbf{250} & 218/250 \\
    & H20 & 2.38 & 63.7 & 1.81 & 1.34 & 1.11 & 247 & \textbf{226}/250 \\
    & L40 & 3.13 & 182 & 1.88 & 1.31 & 1.08 & 248 & 215/250 \\
    \midrule[0.8pt]
    \multirow{5}{*}{\textit{Torch Compile}} 
    & A100 & 2.77 & 69.0 & 2.55 & 1.72 & 1.14 & 249 & 203/250 \\
    & 3090 & 2.58 & 73.2 & 2.23 & 1.50 & 1.00 & 242 & 177/250 \\
    & H100 & 2.74 & 49.7 & 2.83 & 1.92 & 1.11 & \textbf{250} & 195/250 \\
    & H20 & \textbf{2.89} & 49.4 & \textbf{3.21} & \textbf{2.04} & \textbf{1.19} & 247 & \textbf{209}/250 \\
    & L40 & 2.85 & \textbf{96.9} & 2.43 & 1.82 & 1.13 & 248 & 199/250 \\
    \midrule[0.8pt]
    \multirow{5}{*}{\textit{Torch Compile RO}} 
    & A100 & 2.88 & 80.1 & 2.48 & 1.67 & \textbf{1.13} & \textbf{249} & \textbf{200}/250 \\
    & 3090 & 2.61 & 72.9 & 2.29 & 1.48 & 1.00 & 242 & 172/250 \\
    & H100 & 2.77 & 61.2 & 2.78 & 1.61 & 1.00 & 247 & 187/250 \\
    & H20 & 2.82 & 52.1 & \textbf{3.18} & 1.64 & 1.06 & 247 & 192/250 \\
    & L40 & \textbf{2.89} & \textbf{90.9} & 2.54 & \textbf{1.72} & 1.08 & 248 & 193/250 \\
    \midrule[0.8pt]
    \multirow{5}{*}{\textit{CUDA Graph}} 
    & A100 & 2.81 & 97.9 & 1.83 & 1.20 & 0.954 & \textbf{229} & 147/229 \\
    & 3090 & 3.34 & 156 & \textbf{1.94} & \textbf{1.28} & \textbf{0.997} & 206 & \textbf{148/206} \\
    & H100 & 2.23 & 70.1 & 1.60 & 1.04 & 0.838 & 222 & 119/222 \\
    & H20 & 2.20 & 64.6 & 1.69 & 1.09 & 0.854 & \textbf{229} & 133/229 \\
    & L40 & \textbf{3.98} & \textbf{275} & 1.83 & 1.16 & 0.862 & 224 & 137/224 \\
    \bottomrule
    \end{tabular}
    \caption{Performance comparison across different configurations and GPU devices on KernelBench. RO = Reduce Overhead. Speedup is defined as  value exceeding 1.01x.}
    \label{different-gpus-overall}
\end{table*}

 \subsection{Generalization of A100-Optimized Kernels to Other GPU Architectures}
 
Even without being specifically tailored to other GPU architectures, we observe significant performance improvements across all tested GPU types, with mean speedups ranging from 2.38× to 3.85×. H100 achieves the highest mean speedup (3.85×) with exceptional maximum gains (368×), while A100 PCIe and L40 demonstrate strong performance with mean speedups of 3.12× and 3.13× respectively. L40 shows the second-highest maximum speedup (182×) among all GPUs. The consumer RTX 3090 achieves a competitive mean speedup of 2.51×, while H20 shows moderate performance with 2.38× mean speedup. Notably, A100 maintains the highest 75th percentile (2.25×), 50th percentile (1.42×), and 25th percentile (1.17×) values, indicating more consistent optimization performance on the target architecture.

The success rates remain high across all architectures (242-250 out of 250), with H100 achieving perfect success (250/250), validating that CUDA optimization techniques can generalize across different GPU architectures. Speedup achievement rates (>1.01×) vary by architecture, with H20 and A100 showing the highest effectiveness (226 and 226 successful optimizations respectively), while RTX 3090 demonstrates good performance with 201 successful optimizations.

These results demonstrate that while A100-optimized kernels transfer to other GPUs with varying degrees of effectiveness, the optimizations achieve substantial improvements across architectures. H100's exceptional performance suggests strong compatibility with the optimization techniques, while A100's consistent percentile performance validates the target architecture optimization. The varying maximum speedups (63.7× to 368×) across GPUs indicate architecture-specific optimization potential, suggesting that dedicated optimizations for each GPU type would further enhance performance. We plan to release kernels specifically trained for different GPU types in an updated version of CUDA-L1.

\subsection{Discovered Cuda Optimization Techniques}

An analysis of optimization strategies commonly employed in enhanced CUDA implementations reveals interesting patterns. Through GPT-4o-based technical term extraction and frequency analysis, we identified the ten most prevalent optimization techniques:
\begin{itemize}

\item {\bf Memory Layout Optimization}, which ensures data is stored in contiguous memory blocks;
\item  {\bf Memory Access Optimization}, which  arranges data access patterns to maximize memory bandwidth and minimize latency through techniques like shared memory usage, coalesced global memory access, and memory padding;
\item  {\bf Operation Fusion}, which combines multiple sequential operations into a single optimized kernel execution;
\item {\bf Memory Format Optimization}, which  aligns data layout with hardware memory access patterns;
\item  {\bf Memory Coalescing}, which optimizes CUDA kernel performance by ensuring threads in the same warp access contiguous memory locations;
\item   {\bf Warp-Level Optimization}, which leverages the parallel execution of threads within a warp (typically 32 threads) to efficiently perform collective operations;
\item  {\bf Optimized Thread Block Configuration}, which carefully selects grid and block dimensions for CUDA kernels to maximize parallel execution efficiency and memory access patterns;
\item  {\bf Shared Memory Usage}, enables fast data access by storing frequently used data in a cache accessible by all threads within a thread block;
\item  {\bf Register Optimization}, which stores frequently accessed data in fast register memory to reduce latency and improve computational throughput;
\item  {\bf Stream Management}, which enables parallel execution of operations for improved GPU utilization.
\end{itemize}
Tables \ref{optimize-1}, \ref{optimize-2} and \ref{optimize-3}  present detailed CUDA optimization techniques with accompanying code examples.

\section{Conclusion}
latexIn this paper, we propose CUDA-L1, a pipelined system for CUDA optimization powered by contrastive RL. CUDA-L1 achieves significant performance improvements on the CUDA optimization task, delivering an average speedup of ×3.12 (median ×1.42) over the default baseline across all 250 CUDA kernels of KernelBench, with peak speedups reaching ×120 on A100. Against other baselines, CUDA-L1 demonstrates ×2.77 over Torch Compile, ×2.88 over Torch Compile with reduce overhead, and ×2.81 over CUDA Graph implementations. CUDA-L1 can independently discover CUDA optimization techniques, learn to combine them strategically, and more importantly, extend the acquired CUDA reasoning abilities to unseen kernels with meaningful speedups. We hope that CUDA-L1 would open new doors for automated optimization of CUDA, and substantially promote GPU efficiency and alleviate the rising pressure on GPU computing resources.

\newpage 
\appendix








\newpage
\vspace{-3.5cm}
\newpage

\newpage
\bibliography{custom}
\bibliographystyle{acm}
\end{document}